\documentclass[10pt,twocolumn,letterpaper]{article}

\usepackage[pagenumbers]{iccv} 
\usepackage{multirow}
\usepackage{graphicx}
\usepackage{booktabs}
\usepackage{colortbl}
\usepackage[accsupp]{axessibility}
\usepackage{url}
\usepackage{pifont}
\usepackage{marvosym}
\definecolor{iccvblue}{rgb}{0.21,0.49,0.74}
\usepackage[pagebackref,breaklinks,colorlinks,allcolors=iccvblue]{hyperref}

\def\paperID{***} 
\def\confName{ICCV}
\def\confYear{2025}

\title{GUAVA: Generalizable Upper Body 3D Gaussian Avatar}

\makeatletter
\newcommand{\printfnsymbol}[1]{%
  \textsuperscript{\@fnsymbol{#1}}%
}
\renewcommand*{\@fnsymbol}[1]{\ensuremath{\ifcase#1\or *\or \dagger\or \ddagger\or
   \mathsection\or \mathparagraph\or \|\or **\or \dagger\dagger
   \or \ddagger\ddagger \else\@ctrerr\fi}}

\author{
    Dongbin Zhang$^{1,2}$\thanks{Intern at IDEA} \quad Yunfei Liu$^{2}\thanks{Corresponding authors}$ \quad Lijian Lin$^{2}$ \quad Ye Zhu$^{2}$ \quad Yang Li$^{1}$ 
   \\ Minghan Qin$^{1}$ \quad Yu Li$^{2}$\printfnsymbol{3} \quad Haoqian Wang$^{1}$\printfnsymbol{2} \\
    $^1$Tsinghua Shenzhen International Graduate School, Tsinghua University \\ $^2$International Digital Economy Academy (IDEA) 
}

\begin{document}
\twocolumn[{%
\renewcommand\twocolumn[1][]{#1}%
\maketitle
\begin{center}
    \centering
    \captionsetup{type=figure}
    \includegraphics[width=1\textwidth]{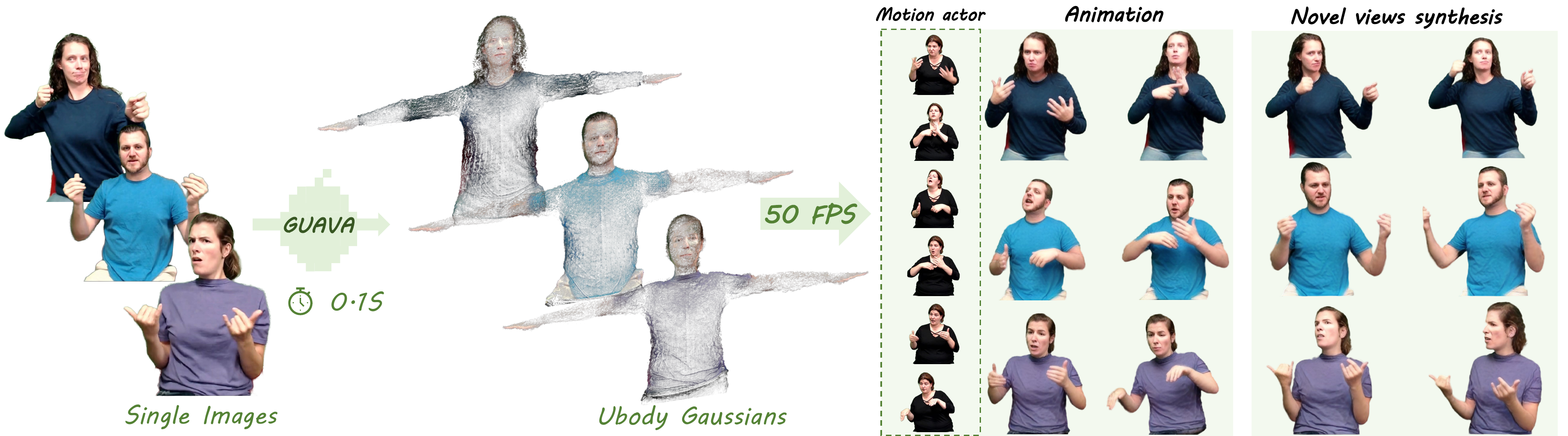}
    \captionof{figure}{From a single image with a tracked pose, GUAVA can reconstruct a 3D upper-body Gaussian avatar via feed-forward inference within sub-second time, enabling real-time expressive animation and novel view synthesis at 512$\times$512 resolution. 
    }
    \label{fig:teaser}
\vspace{-6pt}
\end{center}%
}]

{\let\thefootnote\relax\footnotetext{{$^{*}$ Intern at IDEA. ~ $^{\ddagger}$ Project lead. ~ $^{\dagger}$ Corresponding authors. }}}

\maketitle

\begin{abstract}
Reconstructing a high-quality, animatable 3D human avatar with expressive facial and hand motions from a single image has gained significant attention due to its broad application potential. 3D human avatar reconstruction typically requires multi-view or monocular videos and training on individual IDs, which is both complex and time-consuming. Furthermore, limited by SMPLX’s expressiveness, these methods often focus on body motion but struggle with facial expressions. To address these challenges, we first introduce an expressive human model (EHM) to enhance facial expression capabilities and develop an accurate tracking method. Based on this template model, we propose GUAVA, the first framework for fast animatable upper-body 3D Gaussian avatar reconstruction. We leverage inverse texture mapping and projection sampling techniques to infer Ubody (upper-body) Gaussians from a single image. The rendered images are refined through a neural refiner. Experimental results demonstrate that GUAVA significantly outperforms previous methods in rendering quality and offers significant speed improvements, with reconstruction times in the sub-second range ($\sim$ 0.1s), and supports real-time animation and rendering. Code and video demos are available at the \href{https://eastbeanzhang.github.io/GUAVA/}{\textcolor{magenta}{Project page}}.
\end{abstract}

\section{Introduction}
\label{sec:intro}

Creating realistic and expressive upper-body human avatars is essential for applications like films, games, virtual meetings, and digital media. These avatars are expected to exhibit high realism with expressiveness, such as detailed facial expressions and rich hand gestures. Efficiency, ease of creation, and real-time rendering are also critical. However, achieving these goals, especially from a single image, remains a significant challenge in computer vision. 

Recently, this area has gained growing attention, with progress in both 2D and 3D-based methods. Several works \cite{lin2023motionx,lin2023osx, ExPose:2020, PIXIE:2021, Moon_2022_CVPRW_Hand4Whole} directly predict the coefficients of the human template model SMPLX \cite{SMPL-X:2019} from images, enabling rapid human mesh reconstruction. However, due to the template model's coarse textures, these methods struggle to render photorealistic images. With the breakthrough of NeRF \cite{mildenhall2020nerf} in novel view synthesis, methods like \cite{nerface,zheng2022imavatar,jiang2023instantavatar,ani-nerf,liu2024texvocab} combine template models like FLAME \cite{FLAME:SiggraphAsia2017} or SMPLX to reconstruct high-quality head or whole body avatars from multi-view or monocular videos. While achieving realistic effects, these methods suffer from slow training and rendering speeds. The rise of 3D Gaussian splatting (3DGS) \cite{kerbl3DGS} has led to methods like \cite{hu2024gaussianavatar,qian2024gaussianavatars,shao2024splattingavatar,xiang2024flashavatar,li2024animatablegaussians}, which leverage 3DGS for real-time, high-quality avatars reconstruction. However, these methods still exist several limitations. \eg, \textbf{ID-specific training:} Each person requires individual training; \textbf{Training complexity:} The process is time-consuming and requires calibrated multi-view or monocular videos; \textbf{Limited expressiveness:} Head reconstruction methods lack body motion representation, while full-body methods neglect detailed facial expressions.


In the 2D domain, diffusion-based models \cite{VDM, SVD,chen2023videocrafter1,chen2024videocrafter2} have demonstrated remarkable results in video generation. Meanwhile, ControlNet \cite{controlnet} has expanded the controllability of generative models by adding extra conditions to guide the generation process of stable diffusion model (SD) \cite{SD}. Consequently, some researchers \cite{wang2023disco,karras2023dreampose,hu2023animateanyone} adapt it for human motion transfer. They use conditions like keypoints or SMPLX to transfer the target image's human pose to the source image's ID, enabling the creation of human animation videos. Despite these methods achieving good visual effects, they still face several limitations. \eg, \textbf{ID consistency:} They struggle to maintain a consistent ID without 3D representations, especially with large pose changes; \textbf{Efficiency:} High computational costs and multiple denoising steps result in slow inference, hindering real-time applications; \textbf{Viewpoint control:} 2D methods cannot easily adjust the camera pose, limiting viewpoint control.

To address the above challenges, we propose GUAVA, a framework for creating upper-body 3D Gaussian avatars from single images, as shown in \cref{fig:teaser}. GUAVA enables fast reconstruction of animatable, expressive avatars, supporting real-time pose, expression control, and rendering. Unlike traditional 3D avatar reconstruction methods, GUAVA completes the reconstruction in a single forward pass. Compared to 2D-based methods, we use 3D Gaussians for consistent avatar representation in canonical space, overcoming issues of ID consistency and enabling real-time rendering.


 Like other 3D-based body reconstruction methods, we rely on the human template model SMPLX to construct the upper-body avatar. This requires aligning each image with the template model via tracking \cite{lin2023motionx,Moon_2022_CVPRW_Hand4Whole}. However, current SMPLX parameter estimation methods struggle with accurately tracking hand movements and fine-grained facial expressions. Besides, SMPLX has limited facial expression capability. Therefore, we introduce EHM (Expressive Human Model), combining SMPLX and FLAME, and develop an optimization-based tracking method for accurate parameter estimation. Based on these tracked results, we design a reconstruction model with two branches: one uses the EHM’s vertices and their projection features to create template Gaussians, while the other applies inverse texture mapping to transfer screen-space features into UV space for decoding UV Gaussians. This approach reduces the sparsity of template Gaussians and captures finer texture details. To further improve rendering quality, each Gaussian is equipped with a latent feature to generate a coarse map, which is then refined using a learning-based refiner.
 

 We train GUAVA on monocular upper-body human videos with diverse IDs to ensure good generalization for unseen IDs during inference. Extensive experiments show that GUAVA outperforms previous 2D- and 3D-based methods in visual effects. Additionally, our method reconstructs the upper body in $\sim\mathbf{0.1}s$, and supports real-time animation and rendering. Our main contributions are as follows:
\begin{itemize}
\item  We propose GUAVA, the first framework for generalizable upper-body 3D Gaussian avatar reconstruction from a single image. Using projection sampling and inverse texture mapping, GUAVA enables fast feed-forward inference to reconstruct Ubody Gaussians from the image.
\item We introduce an expressive human template model with a corresponding upper-body tracking framework, providing an accurate prior for reconstruction.
\item Extensive experiments show that GUAVA outperforms existing methods in rendering quality, and significantly outperforms 2D diffusion-based methods in speed, offering fast reconstruction and real-time animation.
\end{itemize}

\section{Related work}
\label{sec:related}

\subsection{3D based Avatar Reconstruction}
Traditional human or head reconstruction primarily focuses on mesh reconstruction. Researchers have constructed human template models like SMPL \cite{SMPL:2015}, the facial model FLAME \cite{FLAME:SiggraphAsia2017}, and the hand model MANO \cite{MANO:SIGGRAPHASIA:2017} based on thousands of 3D scans. These models are mesh-based and represent shape and expression variations in a linear space, with rotational movement modeled using joints and linear blend skinning. SMPLX \cite{SMPL-X:2019} extends SMPL by integrating FLAME and MANO, enhancing facial and hand expressiveness. Since SMPLX is trained on full-body scans and may overlook facial details, its expressive capability for face is still limited compared to FLAME. By combining deep learning, some methods \cite{Goel2023ICCV_HMR2.0,yuan2022glamr, sarandi2025NLF, MobRecon, PointHMR,dong2025hamba, DECA, emoca,tokenface, SMIRK} predict the parameters of human, hand, and face models to achieve fast image-to-mesh reconstruction.

With the development of Neural Radiance Fields (NeRF) \cite{mildenhall2020nerf}, studies have combined NeRF with template models for more realistic 3D reconstructions of heads \cite{zhao2023havatar,athar2022rignerf,qin2024svenerf,INSTA} or humans \cite{li2022tava,weng_humannerf_2022_cvpr,ARAH:2022:ECCV}. NeRFcae \cite{nerface} extends NeRF to a dynamic form by introducing expression and pose parameters as conditioning inputs for driveable avatar reconstruction. Neural Body \cite{neuralbody} proposes a new representation for dynamic humans by encoding posed human meshes into latent code volumes. Recently, 3DGS \cite{kerbl3DGS} has made breakthroughs in real-time rendering, leading to its rapid application in various fields \cite{qin2024langsplat,zhang2024GSW,gao2024relightableGS,2dgs,slgaussian}, leveraging its efficient rendering capabilities, including avatar reconstruction \cite{hu2024gauhuman,qian2024gaussianavatars,codecavatars,liu2024animatable,yuan2024gavatar,hravatar}. GART \cite{lei2024gart} uses 3D Gaussians to represent deformable subjects, employing learnable blend skinning to model non-rigid deformations and generalizing to more complex deformations with novel latent bones.  GaussianAvatar \cite{hu2024gaussianavatar} learns pose-dependent Gaussians using a 2D pose encoder to represent human avatar. ExAvatar \cite{moon2024exavatar} combines mesh and Gaussian representations, treating each Gaussian as a vertex in the mesh with pre-defined connectivity, enhancing facial expression expressiveness.

However, these methods require training for each individual ID and lack generalization. To address this, some works have explored generalizable networks for single image to avatar reconstruction \cite{deng2024portrait4d,portrait4dv2,gpavatar}. GAGAvatar \cite{GAGavatar} combines the FLAME and introduces a dual-lifting method for inferring 3D Gaussians from a single image, enabling feed-forward reconstruction of head avatars. Yet, single-image 3D human reconstruction remains underexplored. Methods like \cite{huang2022elicit, OneTalk,HumanGM,pan2024humansplat} achieve avatar reconstruction from a one-shot image but rely on generating a series of images through generative models, followed by optimization for reconstruction, which is time-consuming. Other methods \cite{enerf, Gps-gaussian,hu2024evags} focus on generalizable models for quick novel view synthesis from sparse images. In this context, we propose GUAVA, to the best of our knowledge, the first method enables fast and generalizable upper-body avatar reconstruction from a single image, while also addressing the lack of body expressiveness in generalizable head avatars.

\subsection{2D based Human Animation}
Early motion transfer frameworks are primarily based on Generative Adversarial Networks (GANs) \cite{gan}, such as those for facial expression transfer \cite{Siarohin_2019_FOMM,face-vid2vid,hong2022depthaware,contrastivegan} and human motion transfer \cite{deformablegan,motiontransfer,warping-gan,wei2020gac,liu2019liquid}. These methods typically achieve motion transfer by warping the source image's feature volume according to the target motion. However, they often generate artifacts in unseen regions and struggle with generalizing to out-of-distribution images. Recently, diffusion models \cite{ddim,scorebased, LDM,instantid, SD} have shown remarkable diversity, high generalization, and superior quality in image generation. By learning from vast image datasets, these models have developed strong priors, enabling them to generalize well to downstream tasks. Some works have integrated pre-trained image generation with temporal layers \cite{makeavideo,tuneavideo,magicvideov2,guo2023animatediff} or transformer structures \cite{yan2021videogpt,yu2023magvit,bao2024vidu,ma2024latte} to achieve video generation. For human motion transfer tasks, these video or image generation models can provide powerful priors. Inspired by these advancements, several works have leveraged generative models to create controllable human animation videos based on pose information extracted from OpenPose \cite{cao2019openpose} or DWPose \cite{dwpose}. For example, DreamPose \cite{karras2023dreampose}, based on Stable Diffusion \cite{SD}, integrates CLIP \cite{clip} and VAE to encode images, using an adapter to generate animation videos from pose sequences. MagicPose \cite{chang2023magicpose}, also an SD-based model, introduces a two-stage training strategy to better decouple human motions and appearance. For more accurate pose control, Champ \cite{zhu2024champ} integrates SMPL as a 3D guide, using depth, normal, and semantic maps as motion control conditions. MimicMotion \cite{mimicmotion2024} employs an image-to-video diffusion model as a prior, proposing a confidence-aware pose guidance to enhance generation quality and inter-frame smoothness, while utilizing a progressive fusion strategy for longer videos generation. Although these 2D-based methods ensure high-quality image synthesis, they all struggle with maintaining ID consistency. In contrast, our method uses 3D Gaussians as the 3D representation for avatars, which ensures better identity consistency.

\begin{figure*}[t]
  \centering
   \includegraphics[width=1.0\linewidth]{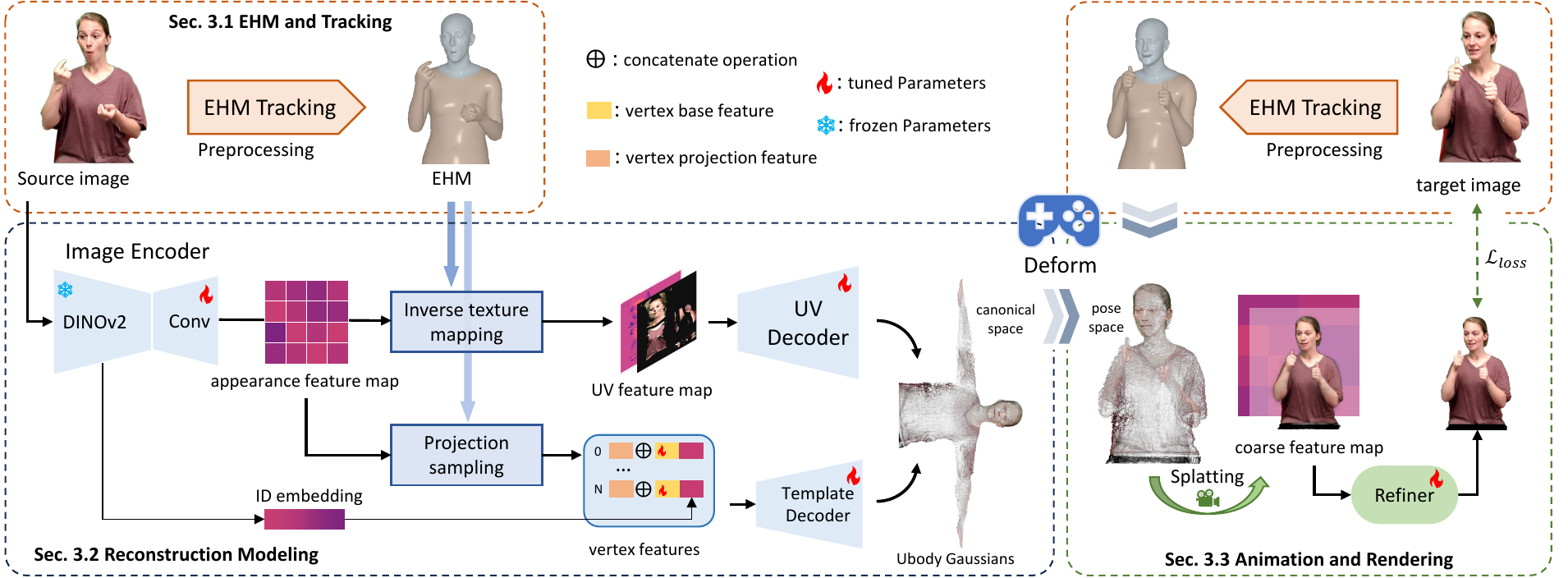}
   \caption{Given the source and target images, we first obtain the shape, expression, and pose parameters of the EHM template model through preprocessing tracking. The source image is then passed through an image encoder to extract an appearance feature map. Using these features and the tracked EHM, one branch predicts the template Gaussians, and the other predicts the UV Gaussians. These are concatenated to form the Ubody Gaussians in canonical space, which are then deformed into pose space using the tracked parameters from the target image. Finally, a coarse feature map is rendered and refined by a neural refiner to produce the final image.}
   \label{fig:pipe}
\end{figure*}

\section{Method}
\label{sec:method}

In \cref{meth:ehm}, we introduce the EHM template model, which enhance SMPLX's \cite{SMPL-X:2019} facial expressiveness, along with an upper-body tracking method. \cref{meth:reco} explains how Ubody Gaussians are predicted from the source image. In \cref{meth:anim}, we deform the Ubody Gaussians into pose space using the tracked parameters and render them. Lastly, \cref{meth:opti} outlines the training strategy and loss functions. The overall pipeline is illustrated in \cref{fig:pipe}.

\subsection{EHM and Tracking}
\label{meth:ehm}
3D body reconstruction often relies on models like SMPLX as priors. Accurately estimating shape and pose parameters to align each frame with SMPLX is essential. However, previous full-body pose estimation methods struggle with robustness and accuracy for wild images, making their results unreliable.  Additionally, while SMPLX struggles to capture fine facial expressions \cite{moon2024exavatar}, FLAME \cite{FLAME:SiggraphAsia2017} excels at this but isn't compatible with SMPLX. This means parameters estimation results based on FLAME cannot be applied to SMPLX. To solve this, we propose EHM (Expressive Human Model), which replaces the SMPLX head part with FLAME for more accurate facial expression representation. Furthermore, we design the tracking in two stages: first, pre-trained models provide a coarse estimate, followed by refinement primarily using 2D keypoint loss.

Specifically, we start by using existing models \cite{liu2025teaser, PIXIE:2021,pavlakos2024reconstructing} to coarsely estimate the shape parameters $\beta_b$ and body pose parameters $\theta_b$ of SMPLX,  hand pose parameters $\theta_h$ of MANO, and facial shape $\beta_f$, expression $\psi_f$, and jaw pose $\theta_{jaw}$ of FLAME. Then, we extract human body keypoints $K_b$ and facial keypoints $K_f$ using keypoint detection models \cite{facealignment,insightface-landmark,lugaresi2019mediapipe,dwpose} for further fine-tuning. We optimize facial parameters of FLAME primarily using keypoint loss: $\beta_f,\psi_f,\theta_{jaw}={\text{argmin}}\  \mathcal{L}_1(\hat{K_f}, K_f)$, where $\hat{K_f}$ are the keypoints from the FLAME. This yields a head model with expression in the neutral pose, denoted as $M_f(\beta_f,\psi_f,\theta_{jaw})$. Next, we replace the head of the SMPLX in the T-pose $M_b(\beta_f)$ with the head model $M_f(\beta_f,\psi_f,\theta_{jaw})$. The modified model then undergoes pose deformation through linear blend skinning ($\mathcal{LBS}$). Additionally, we introduce ID-specific joint offset parameters $\Delta{J}$ for better joint alignment. Thus, the EHM model is formulated as:
\begin{equation}
    \begin{split}
    M_{ehm} = & \mathcal{LBS}\Bigl(\mathcal{U}\bigl(M_b(\beta_b),M_f(\beta_f,\psi_f,\theta_{jaw}) \\
     & + \vec{e}\,\bigr),\theta_b,\theta_h, \mathcal{J}_b(\beta_b)+\Delta{J} \Bigr),
    \end{split}
    \label{eq:ehm}
\end{equation}
where $\mathcal{U}$ represents replacing the SMPLX head with the FLAME, aligned by a displacement vector $\vec{e}$, which represents the displacement between the eye joints of both models, and $\mathcal{J}_b(\beta_b)$ denotes the joints' position. We then fine-tune the parameters mainly using keypoint loss: $\beta_h,~\beta_f,~\theta_b,~\theta_h,~\Delta{J}=\text{argmin}\mathcal{L}_1(\hat{K_b},K_b)$, where $\hat{K_b}$ are the keypoints from the EHM. Notably, our tracking method is designed to optimize multiple frames in parallel to meet the speed requirement.

\subsection{Reconstruction Modeling}
\label{meth:reco}

Unlike person-specific reconstruction methods, our model is designed for single-image feedforward inference, enabling fast upper-body avatar reconstruction. We represent the avatar in a canonical space using Gaussians \cite{kerbl3DGS}, where each Gaussian includes position, rotation, scale, opacity, and a latent appearance feature: $G=\{\mu,r,s,\alpha,c\}$. The avatar consists of two parts: template Gaussians based on EHM, which are fewer in number and handle coarse texture and geometry modeling, and UV Gaussians rigged on the triangulated mesh to capture finer details.

\noindent\textbf{Template Gaussians.}
Specifically, we first extract features from the image using pre-trained DINOV2 \cite{oquab2023dinov2} and obtain a global ID embedding $f_{id}$. Convolutional upsampling is then applied to generate an appearance feature map $F_a$ matching the source image's resolution. Since the person in the source image is posed, we use the tracked EHM in the pose space, project each vertex onto the screen space, and employ linear interpolation $\mathcal{S}$ to sample the projection features:
\begin{equation}
      f_p^{i}=\mathcal{S}\left(F_a,\mathcal{P}\left(v^i,RT_s\right)\right),
      \label{eq:prj}
\end{equation}
where $v^i$ represents the $i\text{-th}$ vertex and ${RT}_s$ is the viewing matrix of the source image. Additionally, for each vertex, we adopt an optimizable base feature $f_b$ to learn unique semantic information. Combining these three features, a Template decoder $\mathcal{D}_{T}$, consisting of MLPs, is used to predict the template Gaussian attributes: $\{r^i,s^i,\alpha^i,c^i\}=\mathcal{D}_{T}(f_p^i\oplus f_b^i\oplus f_{id})$, where $\oplus$ denotes concatenation. For $\mu^i$, we directly take the vertex position $v^i$.

\noindent\textbf{UV Gaussians}.
Using only the EHM's vertices to represent the avatar can not fully capture high-frequency details due to the limited number of vertices, and struggles to represent regions beyond the template model. To address these issues, we propose introducing additional Gaussians to capture fine details. Specifically, we use the UV texture map to construct UV Gaussians, where we predict a Gaussian for each valid pixel in the texture map. Unlike template Gaussians, which are directly represented in world coordinates, as in \cite{qian2024gaussianavatars}, we rig each UV Gaussian to the corresponding mesh triangle. Each UV Gaussian is defined as $G_{uv}=\{\Delta{\mu},r,s,\alpha,c,k,t\}$, where $\Delta{\mu}$ represents the local position of the Gaussian in the triangle, $k$ is the triangle index, and $t$ refers to interpolated position on the triangle using the pixel's barycentric coordinates. During rendering, these properties are transformed into world coordinates:
\begin{equation}
    r'=R_tr,~\mu'=\sigma R_t\Delta\mu+t,~s'=\sigma s,
    \label{eq:2world}
\end{equation}
where $\sigma$ and $R_t$ represent the average edge length and orientation of the triangle in world coordinates. Rigging the Gaussians in the local coordinates of the triangle, they can flexibly model parts that the EHM template model cannot represent. Additionally, the increased number of Gaussians enhances the model's ability to capture finer details.


\noindent\textbf{Inverse texture mapping.}
To effectively predict Gaussian attributes in the UV space based on the source image, we design an inverse texture mapping process. This explicitly maps the appearance feature map from screen space to the UV texture space, resulting in a UV feature map $F_{uv}$. Utilizing the tracked EHM mesh, we first map each pixel of the UV map to the mesh by interpolating the triangle’s vertex positions with pixel's barycentric coordinates.  Next, each mapped position $t$ is projected into screen space using camera parameters, followed by linear sampling on the appearance feature map. These sampled features are then placed back into the UV space. Finally, we use a mesh rasterizer to retrieve the visible pixel region and filter out the invisible ones from the UV features.
After obtaining the UV feature map, we input it into the UV decoder to predict the corresponding Gaussian attributes: $\{\Delta{\mu},r,s,\alpha,c\}=\mathcal{D}_{UV}(F_{uv})$. The UV decoder consists of StyleUNet and convolutional networks. First, StyleUNet paints the invisible regions, then the convolutional networks predict each Gaussian attribute.

\subsection{Animation and Rendering}
\label{meth:anim}
\textbf{Animation.}
After reconstructing the upper body avatar in the canonical space, we can animate the reconstructed avatar with the new tracking parameters to present new expressions and motions. The template Gaussians inherit their positions from the EHM vertices, so we use the deformed EHM vertex positions (\cref{eq:ehm}) as the template Gaussian positions in the pose space.  Other Gaussian attributes remain the same, except for the rotation, which undergoes a consistency rotation: $r'=R_{lbs}r$, where $R_{lbs}$ represents the weighted blend skinning rotation matrix. For the UV Gaussians, each Gaussian remains fixed in the local coordinate system relative to its rigged parent triangle, but moves in world coordinates as the triangle deforms. During animation, after obtaining the deformed EHM mesh, we compute the orientation $R_t$ and average edge length $\sigma$ for each triangle. Then, we use \cref{eq:2world} to transform the UV Gaussians into world coordinates.

\noindent\textbf{Rendering.}
Although we increase the number of Gaussians by adding UV Gaussians, the number of valid Gaussians after reconstruction may be less than 150000. For complex upper body models, the sparsity of Gaussians may limit the expressiveness. To address this and enhance rendering quality, following \cite{li2024animatablegaussians,ultraGSavatar}, we assign each Gaussian with a latent feature $c$. During rendering, we first obtain a coarse feature map $I_F$ through splatting where the first three dimensions represent the coarse RGB image $I_c$. Then, the image features are passed through a StyleUNet-based refiner~\cite{GAGavatar}, which decodes them into a refined image $I_r$ of the same resolution. Compared to direct RGB rendering, this method strengthens the implicit representation of the Gaussian and improves detail capture.


\subsection{Training Strategy and Losses}
\label{meth:opti}
Similar to the training strategy of \cite{drobyshev2022megaportraits,gpavatar}, we randomly select two images from each image sequence: one as the source image to reconstruct the avatar and the other as the target image to drive the avatar. The loss between the driven result and the target image is then used to train the model.  For image loss, we use $\mathcal{L}_1$ and perceptual loss LPIPS $\mathcal{L}_{lpips}$ \cite{LPIPS}. To enhance the model's focus on local details, especially the face and hands, we also crop these regions from the image and include them in the loss calculation:
\begin{equation}
    \begin{split}
    \mathcal{L}_{image}=&\mathcal{L}_c(I_t,I_r)+\mathcal{L}_c(I_t,I_c)+\lambda_f\cdot\mathcal{L}_c(I_t^{face},\\&I_r^{face})
    +\lambda_h\cdot\mathcal{L}_c(I_t^{hand},I_r^{hand}),~\text{where}\\
    \mathcal{L}_c(I_t,I_r)&=\lambda_1\cdot\mathcal{L}_1(I_t,I_r)+\lambda_{lpips}\cdot\mathcal{L}_{lpips}(I_t,I_r).
    \end{split}
\end{equation}
Following \cite{qian2024gaussianavatars}, we introduce a regularization loss to ensure that the Gaussians rigged on the mesh stay close to the thier parent triangles: $\mathcal{L}_{pos}=||\text{max}(\Delta\mu,\epsilon_{pos})||_2$, where $\epsilon_{pos}$ allowed offset threshold relative to the parent triangle's scale. We also regularize the scaling of UV Gaussians: $\mathcal{L}_{sca}=||\text{max}(s,\epsilon_{sca})||_2$, where $\epsilon_{sca}$ represents the allowed scaling threshold.  The total loss function is shown as \cref{eq:tloss}, where $\lambda$ denotes the weight coefficient.
\begin{equation}
\mathcal{L}=\mathcal{L}_{image}+\lambda_p\cdot\mathcal{L}_{pos}+\lambda_s\cdot\mathcal{L}_{sca}.    
\label{eq:tloss}
\end{equation}

\begin{figure*}[t]
  \centering
   \includegraphics[width=0.95\linewidth]{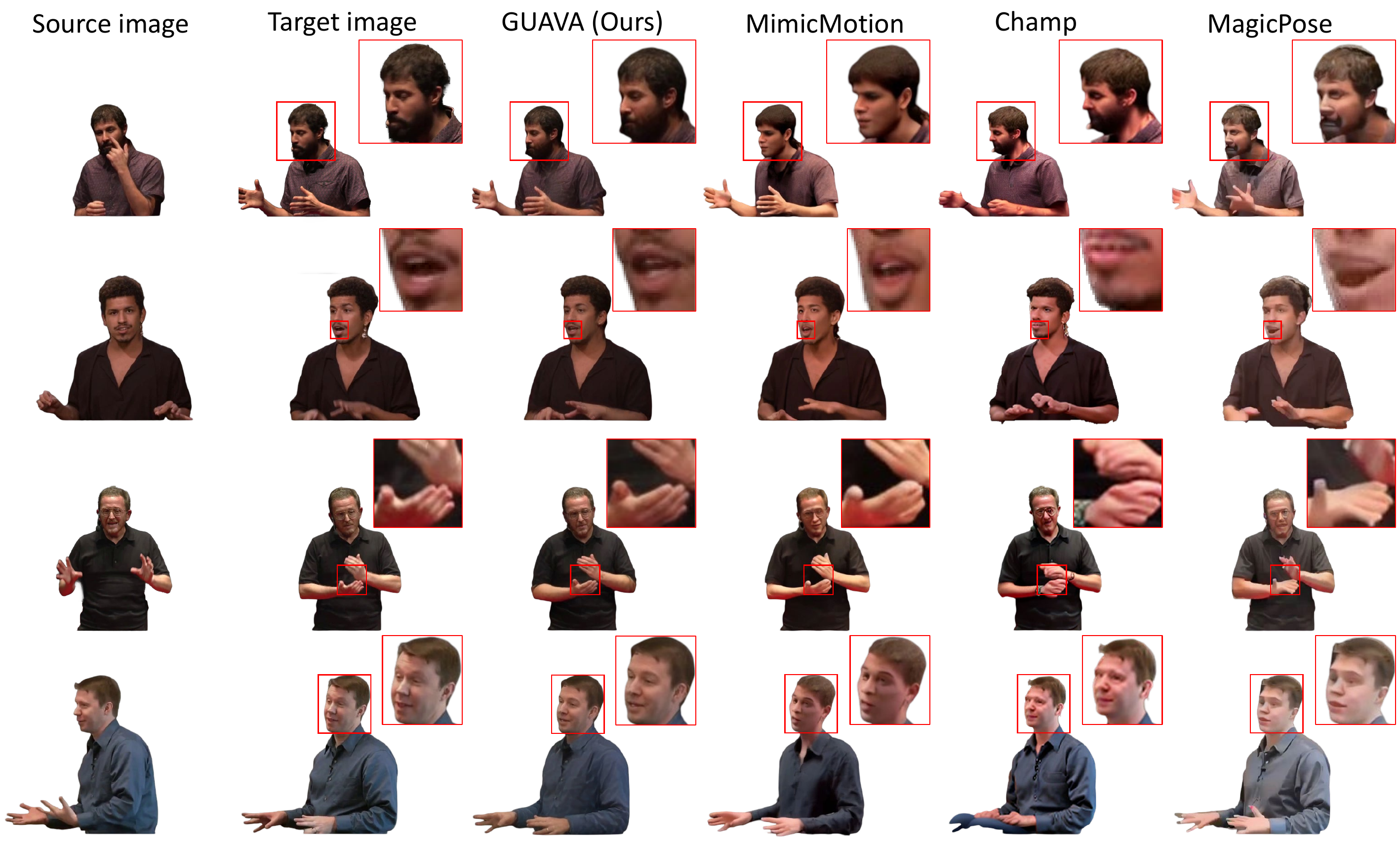}
   \caption{Qualitative comparison results on self-reenactment. Compared to others, our method better preserves ID consistency during animation while capturing more detailed facial expressions and hand gestures.}
   \label{fig:2d-self-act}
\end{figure*}

\begin{figure*}[t]
  \centering
   \includegraphics[width=0.95\linewidth]{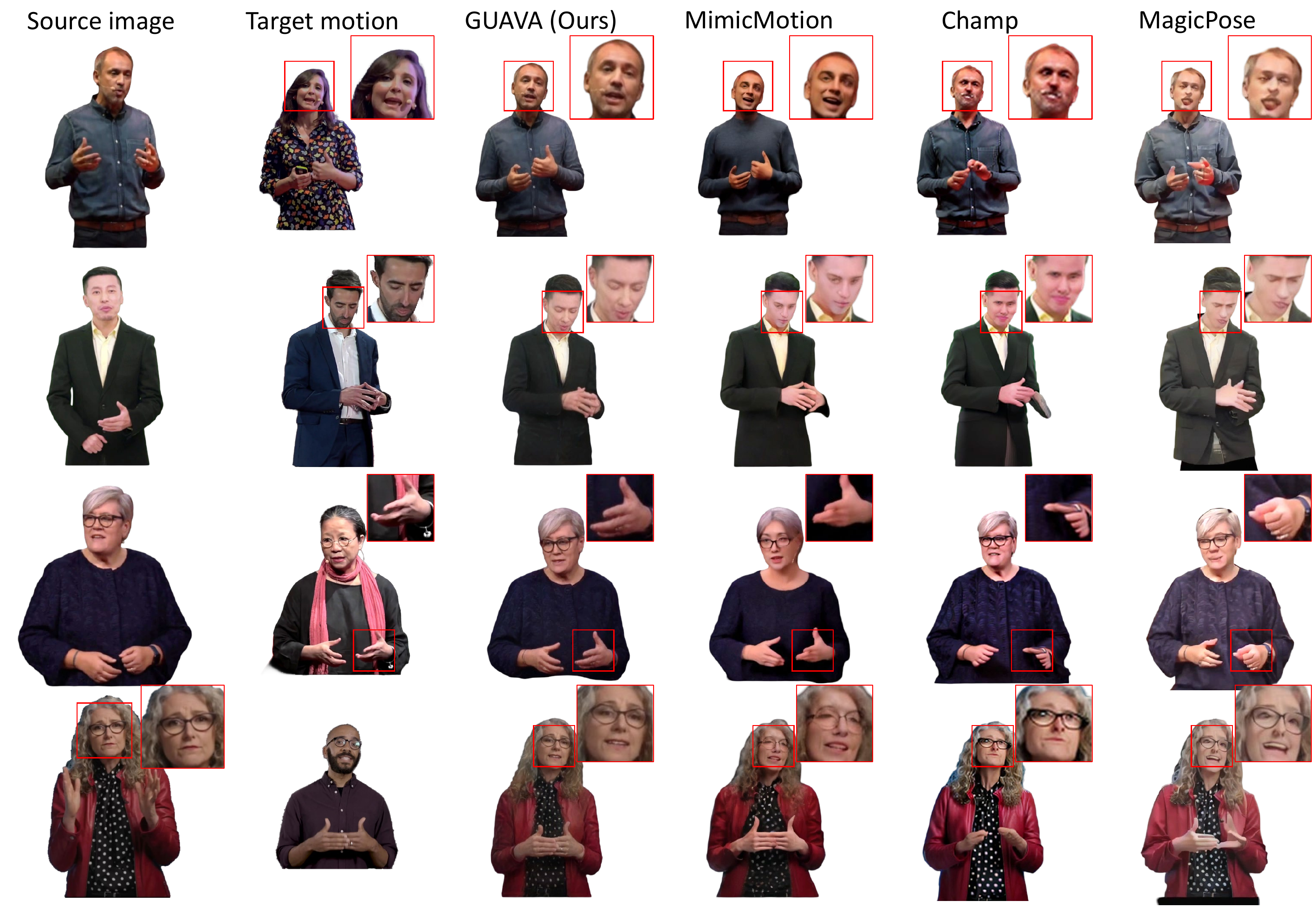}
   \caption{Qualitative results on cross-reenactment against 2D-based methods. Our method demonstrates superior performance in preserving ID consistency across various poses, as well as more accurately capturing the facial expressions and hand gestures of the target motion.}
   \label{fig:2d-cross-act}
   \vspace{-2mm}
\end{figure*}

\section{Experiment}
\label{sec:expriment}

\subsection{Experimental Setup}
\textbf{Implementation details.} Our model is built with PyTorch \cite{paszke2019pytorch} and trained using the Adam optimizer \cite{kingma2014adam}. We train the model on NVIDIA RTX A6000 GPUs for 200,000 iterations, consuming about 156 GPU hours, with a total batch size of 12. For data processing, we use PIXIE \cite{PIXIE:2021}, TEASER \cite{liu2025teaser}, and HaMeR \cite{hamer} to perform rough estimations of SMPLX, FLAME, and MANO parameters, respectively. Body masks are extracted using \citep{stylematte}, and the background is set to black. Moreover, we extract human and facial keypoints using \cite{dwpose,lugaresi2019mediapipe,facealignment,insightface-landmark}.


\noindent\textbf{Dataset.} We collect a dataset consisting of videos from YouTube, OSX \cite{lin2023osx}, and HowToSign \cite{how2sign}, with a focus on human upper body videos. During preprocessing, 5 to 75 frames are sampled at equal intervals from each video depending on the video length. The final training set contains over 26,000 video clips and 620,511 frames. The frames with low hand keypoints confidence due to occlusion are discarded. For the test set, 58 randomly selected IDs are used, with one video per ID, each averaging 15 seconds, totaling 28,287 frames.

\noindent\textbf{Metrics.} \textit{Self-reenactment.} We use the first frame of each video as the source image and generate animations based on the motion from the entire video. We evaluate image quality between animated results and original video using PSNR, L1, SSIM, and LPIPS. \textit{Cross-reenactment.} When driving the current ID with motion from another video, no ground truth exists. We use ArcFace \cite{deng2018arcface} to measure the Identity Preservation Score (IPS) which computes the cosine similarity of identity features.

\noindent \textbf{Baseline.} \textit{2D-based methods.} We compare GUAVA with several SOTA methods: MagicPose \cite{chang2023magicpose}, Champ \cite{zhu2024champ}, and MimicMotion \cite{mimicmotion2024}, which enable controllable human animation video synthesis from a single image. \textit{3D-based methods.} We also compare GUAVA with SOTA 3D-based methods like GART \cite{lei2024gart}, GaussianAvatar \cite{hu2024gaussianavatar}, and ExAvatar \cite{moon2024exavatar}. These methods reconstruct 3D human avatars from monocular videos. For comparison, we use the first half of each video for their training and the remaining half for self-reenactment evaluation, while GUAVA uses the first frame as the source image.

\begin{table}[t]
\centering
\resizebox{0.95\linewidth}{!}{
\begin{tabular}{ccccc|c}
     \toprule
       &  PSNR$\uparrow$ &  $\text{$\mathcal{L}_1$}$$\downarrow$ & SSIM$\uparrow$ & LPIPS$\downarrow$ & FPS$\uparrow$  \\

    \midrule
   GUAVA (Ours)   & \bf 25.87 & \bf0.0162 & \bf0.9000 & \bf0.0813 &\bf 52.21 \\
  MimicMotion   & 24.46 & 0.0200 & 0.8768 & 0.0879  &0.21\\ 
  Champ   &22.01  &0.0258  &0.8643 &0.1000   &0.53  \\
   MagicPose  & 21.25 & 0.0333 & 0.8661 & 0.0913 &0.12   \\

   \bottomrule
\end{tabular}
}
\caption{ Quantitative results on the self-reenactment against 2D-based methods. FPS denotes the animation and rendering speed.}
\label{tb:self-act-2d}
\end{table}

\begin{table}[t]
\centering
\resizebox{0.95\linewidth}{!}{
\begin{tabular}{ccccc|cc}
     \toprule
       &  PSNR$\uparrow$ &  $\text{$\mathcal{L}_1$}$$\downarrow$ & SSIM$\uparrow$ & LPIPS$\downarrow$ & Recon. input & Recon. time\\

    \midrule
   GUAVA (Ours)   & \bf 25.70 & \bf0.0168 & \bf0.8976 & \bf0.0836 & first frame &\bf $\approx$ 98 ms \\
  ExAvatar   & 24.09 & 0.0207 & 0.8783 & 0.1064   & half video &$\approx$ 2.4 h \\ 
  GaussianAvatar & 23.62  & 0.0199 & 0.8780 & 0.1085   &half video  &$\approx$ 1.3 h\\
   GART  & 24.46 & 0.0195 & 0.8805 & 0.1016 &half video &$\approx$ 7 min  \\

   \bottomrule
\end{tabular}
}
\caption{ Quantitative results on the self-reenactment against 3D-based methods. Recon. denotes reconstruction.}
\label{tb:self-act-3d}
\end{table}

\begin{table}[t]
\resizebox{1.0\linewidth}{!}{
\begin{tabular}{ccccc}
     \toprule
       &  GUAVA (Ours) & MimicMotion& Champ & MagicPose   \\

    \midrule
   IPS$\uparrow$  & \bf 0.5554 & 0.1310 & 0.3677 & 0.3277 \\

   \bottomrule
\end{tabular}
}
\caption{ ID Preservation Score on cross-reenactment against 2D-based methods. Our method maintains consistent identity.}
\label{tb:cross-act-2d}
\end{table}

\subsection{Evaluation}
\noindent \textbf{Quantitative results.} \textit{Self-reenactment.} For 2D-based methods, we evaluate all video frames, while for 3D-based methods, we assess the latter half. As shown in \cref{tb:self-act-2d} and \cref{tb:self-act-3d}, our method outperforms all others across all metrics, demonstrating high-quality avatar reconstruction with accurate motion and photorealistic rendering.
\textit{Cross-reenactment.} We use 10 source images from the test set and ViCo-X \cite{vicox}, driven by 8 videos (5079 frames). \cref{tb:cross-act-2d} shows that our method significantly outperforms others in IPS, proving its superior ability to preserve identity consistency with the source ID under different poses.
\textit{Efficiency.} We evaluate all methods on an NVIDIA RTX 3090 GPU, with results shown in \cref{tb:self-act-2d} and \cref{tb:self-act-3d}. Our method achieves around 50 FPS in animation and rendering, while other 2D-based methods take several seconds per frame. 3D-based methods also support real-time rendering but require minutes to hours for reconstruction, whereas our method completes it from a tracked image in just 0.1s.

\begin{figure}[t]
  \centering
   \includegraphics[width=1.0\linewidth]{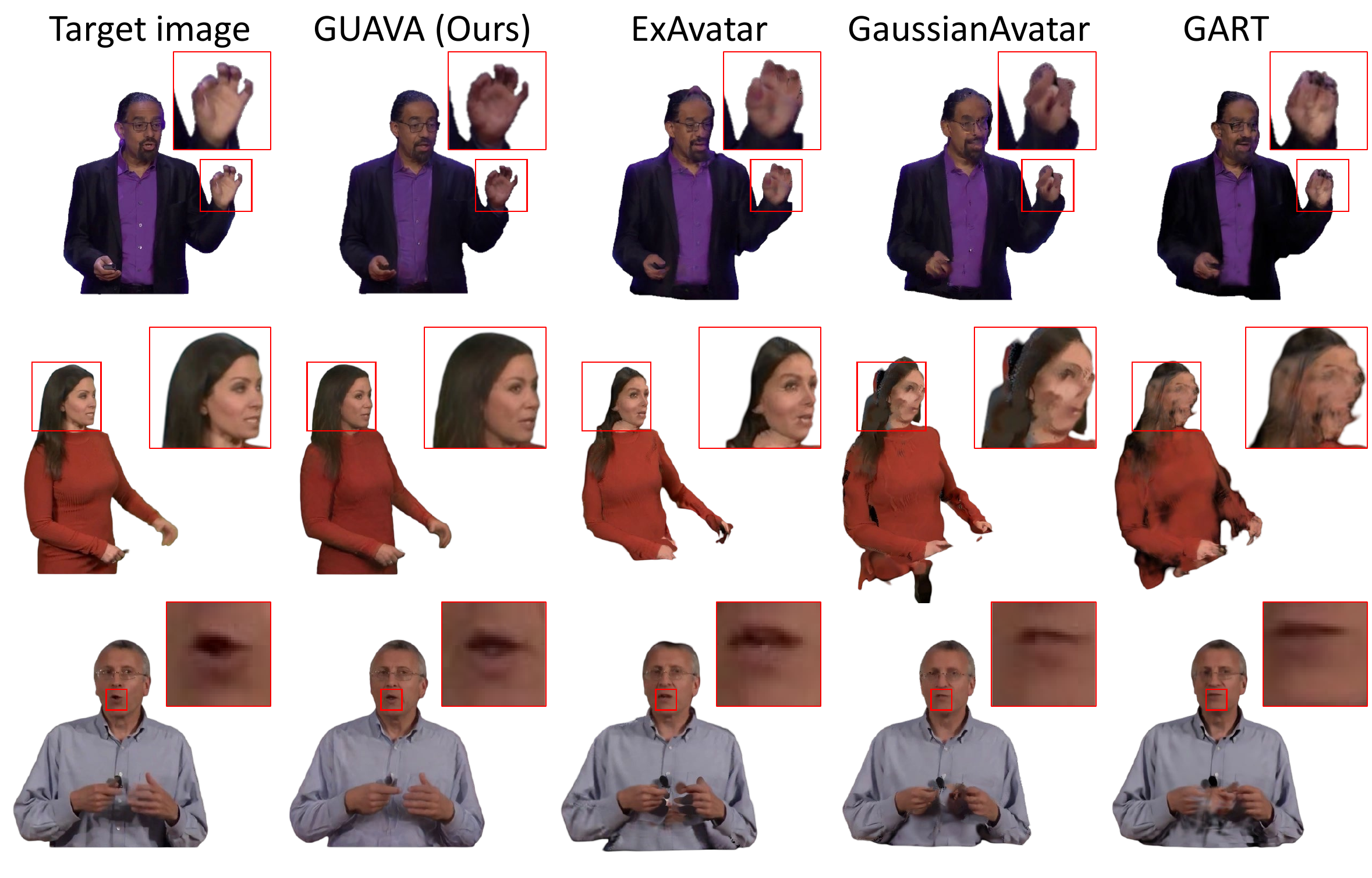}
   \caption{Visual results on self-reenactment against 3D-based methods. Our method reasonably generates unseen regions while capturing more detailed facial expressions and hand gestures.}
   \label{fig:3d-self-act}
\end{figure}

\noindent \textbf{Qualitative results.} \textit{2D-based methods.} The visual comparison results for self-reenactment and cross-reenactment are shown in \cref{fig:2d-self-act} and \cref{fig:2d-cross-act}. Leveraging strong diffusion model priors, 2D-based methods can generate high-quality images. However, the animations from Champ struggle to accurately recover gestures and facial expressions, with blurred hands. MagicPose faces similar issues, with noticeable color distortions. MicmicMotion exhibits better gesture and facial expression performance but fails to maintain identity consistency. In contrast, our approach not only maintains identity consistency with the source image but also restores complex gestures, head poses, and detailed facial expressions like blinking and talking.  
\textit{3D-based methods.} The qualitative results under the self-reenactment setting are shown in \cref{fig:3d-self-act}. GaussianAvatar and GART struggle with fine finger and facial expression driving. ExAvatar performs better in these areas, but these methods lack generalization, producing incomplete results for unseen regions and large artifacts in extreme poses. In contrast, our method generates reasonable results for unseen areas, shows better robustness in extreme poses, and provides more accurate and detailed hand and facial expressions.

\begin{table}[t]
\centering
\resizebox{0.95\linewidth}{!}{
\begin{tabular}{ccccc}
     \toprule
       &  PSNR$\uparrow$ &  $\text{$\mathcal{L}_1$}$$\downarrow$ & SSIM$\uparrow$ & LPIPS$\downarrow$   \\

    \midrule
   Full (Ours)   & \bf 25.87 & \bf0.0162 & \bf0.9000 & \bf0.0813  \\
  w/o refiner   & 24.93 & 0.0188 & 0.8851 & 0.1060  \\ 
  w/o inverse   &25.65  &0.0168  &0.8977 &0.0864     \\
   w/o UV Gaussians  & 25.82 & 0.0164 & 0.8971 & 0.0877 \\
   w/o EHM  & 25.60 & 0.0168 & 0.8950 & 0.0846    \\

   \bottomrule
\end{tabular}
}
\caption{ Ablation results under the self-reenactment setting.}
\label{tb:ablation}
\end{table}

\begin{figure}[t]
  \centering
   \includegraphics[width=1.0\linewidth]{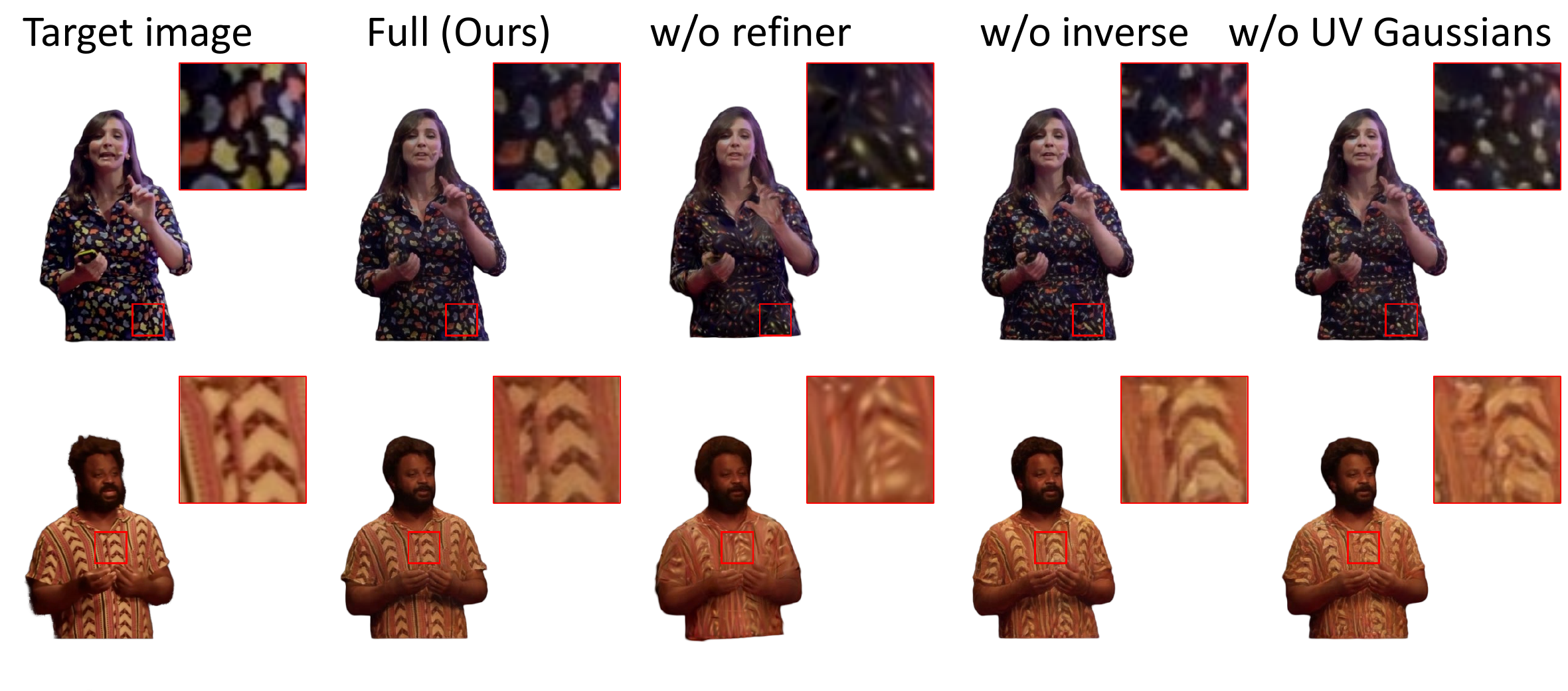}
   \caption{Qualitative results of ablation. Our full method more accurately captures texture details from the source image.}
   \label{fig:ablation-texture}
\end{figure}

\begin{figure}[t]
  \centering
   \includegraphics[width=\linewidth]{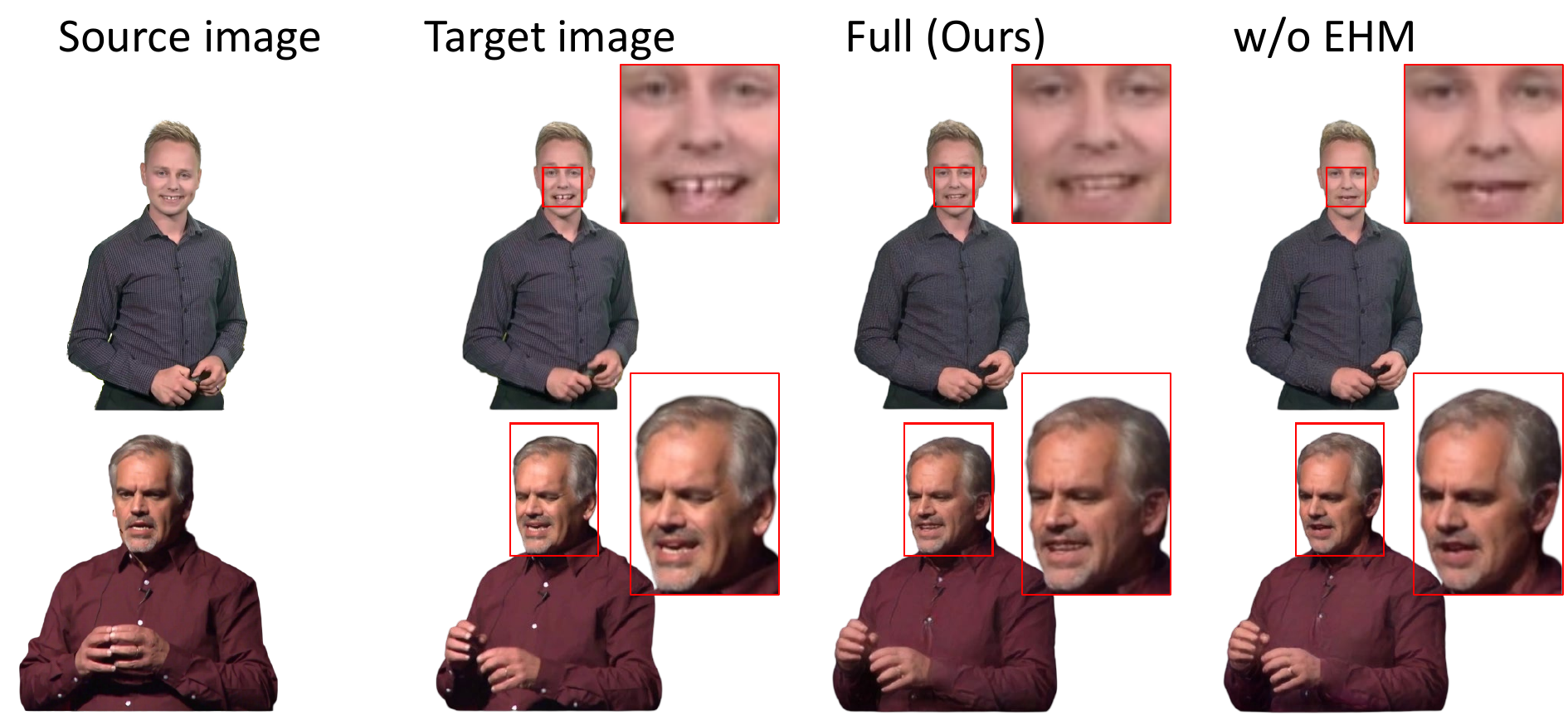}
   \caption{Ablation results of EHM. Using EHM, our method more finely recovers facial expressions and shapes.}
   \label{fig:ablation-ehm}
\end{figure}

\subsection{Ablation Studies}
\label{exp:ablation}
To validate each component's effectiveness, we perform ablation studies under the self-reenactment setting. The quantitative results are summarized in \cref{tb:ablation}.

\noindent \textbf{w/o refiner.} Without the refiner, sparse Gaussians struggle to capture high-frequency texture details accurately. As shown in \cref{fig:ablation-texture}, this leads to elliptical artifacts when dealing with intricate patterns. The significant drop in the LPIPS metric further confirms this limitation.

\noindent \textbf{w/o inverse.} This setting removes inverse UV mapping during UV Gaussians synthesis, directly feeding the screen-space appearance feature map into the UV decoder. While the StyleUnet in the UV decoder can implicitly learn some mappings, inaccuracies occur. As shown in \cref{fig:ablation-texture}, it struggles to capture structured textures precisely, resulting in blurred outputs and lower performance across all metrics.

\noindent \textbf{w/o UV Gaussians.} Relying solely on a limited number of template Gaussians, the model maintains a comparable PSNR but lacks expressiveness, leading to blurry outputs, as reflected in the lower LPIPS score. \cref{fig:ablation-texture} further illustrates this issue, where the model struggles to reproduce high-frequency clothing textures.

\noindent \textbf{w/o EHM.} Using the original SMPLX for tracking and reconstruction weakens the model’s ability to capture fine facial details. As shown in \cref{fig:ablation-ehm}, it sometimes struggles to track facial expressions or head shapes precisely, causing reconstruction and driving errors. The overall performance also drops across all metrics.
\section{Discussion}
\label{sec:discussion}

\noindent \textbf{Conclusion.} We propose GUAVA, a fast framework for expressive 3D upper-body Gaussian avatar reconstruction from a single image, enabling real-time animation and rendering. To enhance facial expression, shape, and pose tracking, we introduce EHM with an accurate tracking method. We further propose inverse texture mapping and projection sampling to reconstruct Ubody Gaussians, composed of UV Gaussians and template Gaussians, where UV Gaussians enhance texture details. Extensive experiments show that GUAVA provides accurate animation, high-quality rendering, and superior efficiency.

\noindent \textbf{Limitation.} Although GUAVA achieves high-quality rendering, its geometric accuracy is limited by training data diversity. Details like clothing wrinkles, loose garments, and hair are difficult to reconstruct. Additionally, limited by the mesh prior, GUAVA struggles to capture regions far from the mesh, making large hairstyles like afros difficult to reconstruct. The lack of clothing deformation modeling limits realistic pose-driven garment changes. Finally, since the dataset focuses on frontal views, GUAVA cannot generate a full 360-degree avatar. These limitations suggest areas for future research and improvements.

{
    \small
    \bibliographystyle{ieeenat_fullname}
    \bibliography{main}
}










\maketitlesupplementary
\thispagestyle{empty}
\appendix

\section*{Overview}
\noindent This supplementary material presents more details and additional results not included in the main paper due to page limitation. The list of items included is:
\begin{itemize}
\item Video demo at \href{https://eastbeanzhang.github.io/GUAVA/}{\textcolor{magenta}{Project page}} with a brief description in \cref{spsec:videodemo}.
\item More model implementation details in \cref{spsec:more-im-de}.

\item More results and EHM tracking in \cref{spsec:more-result}.
\item Further discussion and ethical considerations in \cref{spsec:more-discuss}.

\end{itemize}

\section{Video Demo}
\label{spsec:videodemo}
We highly recommend readers watch the video demo in the supplementary materials. The video showcases GUAVA’s self-reenactment and cross-reenactment animation results, as well as novel view synthesis. Additionally, we compare GUAVA with both 2D-based \cite{moon2024exavatar,lei2024gart,hu2024gaussianavatar} and 3D-based \cite{mimicmotion2024,zhu2024champ,chang2023magicpose} methods under self-reenactment. We also compare it with MimicMotion \cite{mimicmotion2024}, Champ \cite{zhu2024champ}, and MagicPose \cite{chang2023magicpose} under cross-reenactment. Finally, we present the visual results of ablation studies. These results demonstrate that our method enables more detailed and expressive facial and hand motion while maintaining ID consistency with the source image across various poses.

\begin{figure*}[t]
  \centering
   \includegraphics[width=1.0\linewidth]{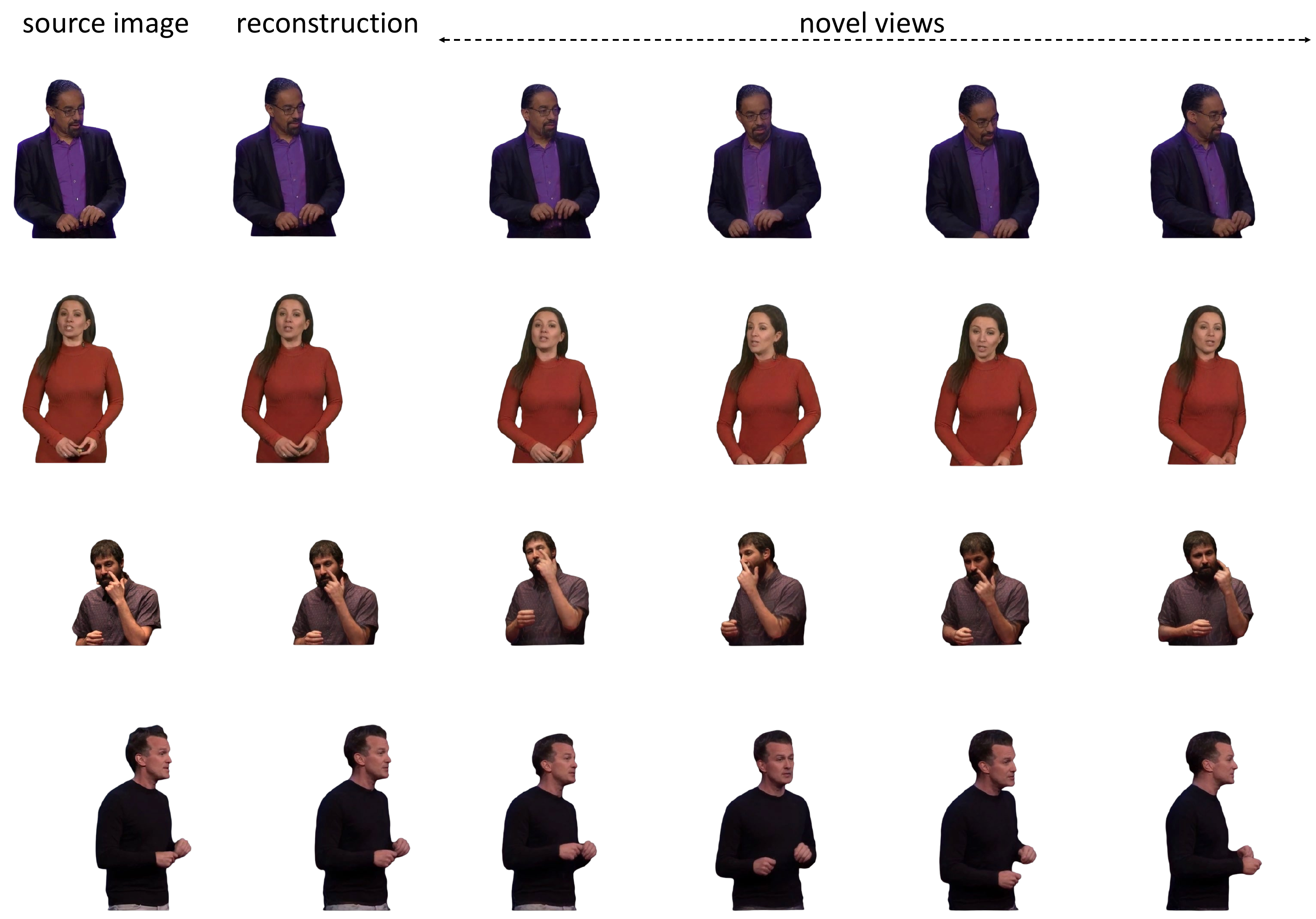}
   \caption{Visual results on novel view synthesis. Our method effectively generates reasonable 3D information while ensuring strong multi-view consistency and preserving the details of the source image.}
   \label{supfig:show-novel-views}
\end{figure*}

\section{More Implementation Details}
\label{spsec:more-im-de}
\subsection{Training details}
We train the model for a total of 200,000 iterations with a fixed learning rate of 1e-4. The learning rate for certain MLPs gradually decreases linearly to 1e-5 over the training process, while the weights of DINOv2 \cite{oquab2023dinov2} remain frozen. Initially, we set the LPIPS \cite{LPIPS} loss weight $\lambda_{lpips}$ to 0.025 and increase it to 0.05 after 10,000 iterations. Other loss weights are set as follows: $\lambda_{f}=0.25$, $\lambda_{h}=0.1$, $\lambda_{1}=1.0$, $\lambda_{p}=0.01$, $\lambda_{s}=1.0$. The hyperparameters $\epsilon_{pos}$, $\epsilon_{sca}$ are set to 3.0 and 0.6, respectively.

\subsection{Model details}
In the reconstruction model, the appearance feature map $F_{a}$ output by the image encoder is passed through convolutional layers to transform its dimensions to 32 and 128, which are then used for the UV Gaussians prediction and template Gaussians prediction branches, respectively.

In the UV branch, the appearance feature map is concatenated with the original image, and inverse texture mapping is applied to map the features to the UV space, resulting in $F_{uv}\in\mathbb{R}^{H\times W\times 35}$. This is passed into the UV decoder’s StyleUnet, which outputs a 96-dimensional feature map. The feature map is then further processed by a convolutional module to decode the Gaussian attributes for each pixel. Additionally, the ID embedding $f_{id}$ is injected into StyleUnet via an MLP.

In the template Gaussian prediction branch, the projection feature $f_p$ and the base vertex feature $f_b$ are both set to a dimension of 128. The ID embedding $f_{id}$ is mapped to 256 dimensions via an MLP.

For Gaussian representation, we discard the spherical harmonic coefficients and use a latent feature $c$ with a dimension of 32 to model the Gaussian appearance. Through splatting, we obtain a rough feature map with a dimension of 32. To help the refiner decode finer images from the rough feature map, we use a loss function to ensure that the first three channels of the latent feature represent RGB. Some details of the model are not shown in Fig.~2 of the main paper, for clarity.

\subsection{Evaluation details}
For self-reenactment evaluation, MagicPose struggles with synthesizing black backgrounds. To avoid the background color influencing the evaluation metrics, we use the ground truth mask to remove the background. Similarly, for Champ, since it uses SMPL-rendered maps \cite{SMPL:2015} (\eg, normal and depth) as input, the generated images may include legs. To ensure accurate metric calculations, we also apply the ground truth mask to filter out the irrelevant parts.

\subsection{Inverse texture mapping}
Here, we explain inverse texture mapping with added details for clarity and ease of understanding. Given the tracked mesh and its corresponding information, including the vertices of each triangle and the three UV coordinates for each triangle, we can locate the area covered by each triangle on the UV map. Then, we identify which triangle each pixel belongs to and calculate its barycentric coordinates. For each pixel, we use its barycentric coordinates to interpolate the triangle vertices and calculate the corresponding position $t$ on the mesh. Next, we project each pixel onto screen space based on its position $t$:

\begin{equation}
x_{uv}^j=\mathcal{P}(t^j,RT_s),~j\in[0,H\cdot W],    
\label{speq:inverpj}
\end{equation}
where $RT_s$ is the viewing matrix of the source image and $\mathcal{P}$ denotes projection. Finally, we perform linear sampling on the appearance feature map, reshape it to $H\times W\times 35$, and obtain $F_{uv}$, completing the inverse texture mapping of the appearance feature map to UV space. To filter out features from invisible regions, we use the \href{https://pytorch3d.org/}{\textcolor{cyan}{Pytorch3D}} rasterizer to render the tracked mesh and acquire the visible triangles. For pixels corresponding to invisible triangles, their features are set to 0.

\begin{figure*}[t]
  \centering
   \includegraphics[width=1.0\linewidth]{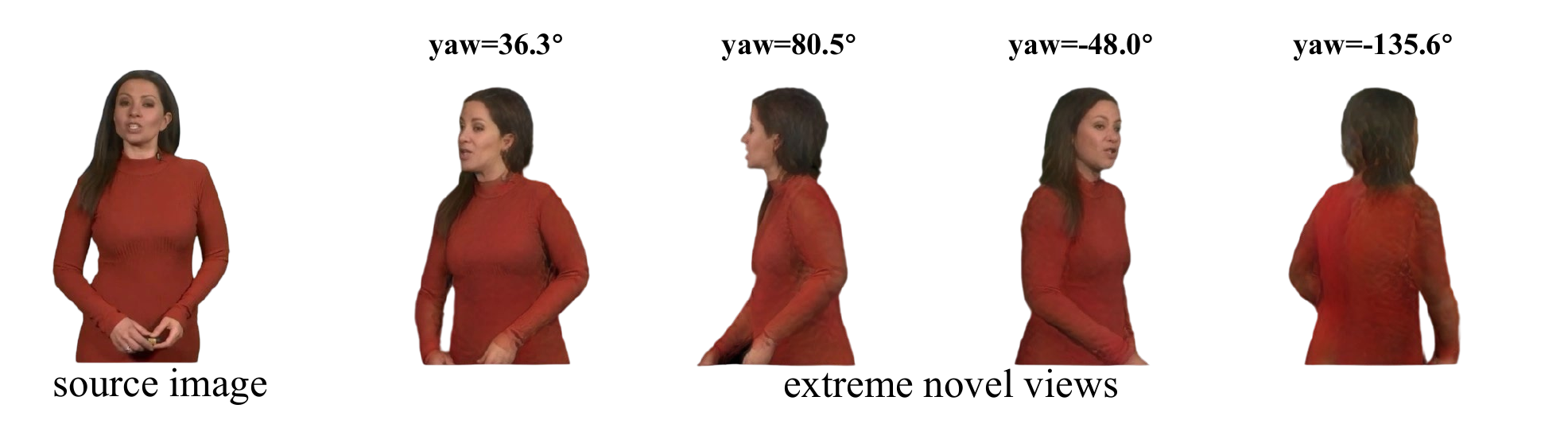}
   \caption{Visual results on extrapolated novel view synthesis. Renderings of back-facing regions are slightly lower quality due to the lack of backside data in our training set.}
   \label{supfig:show-extreme-novel}
\end{figure*}

\begin{figure*}[t]
  \centering
   \includegraphics[width=1.0\linewidth]{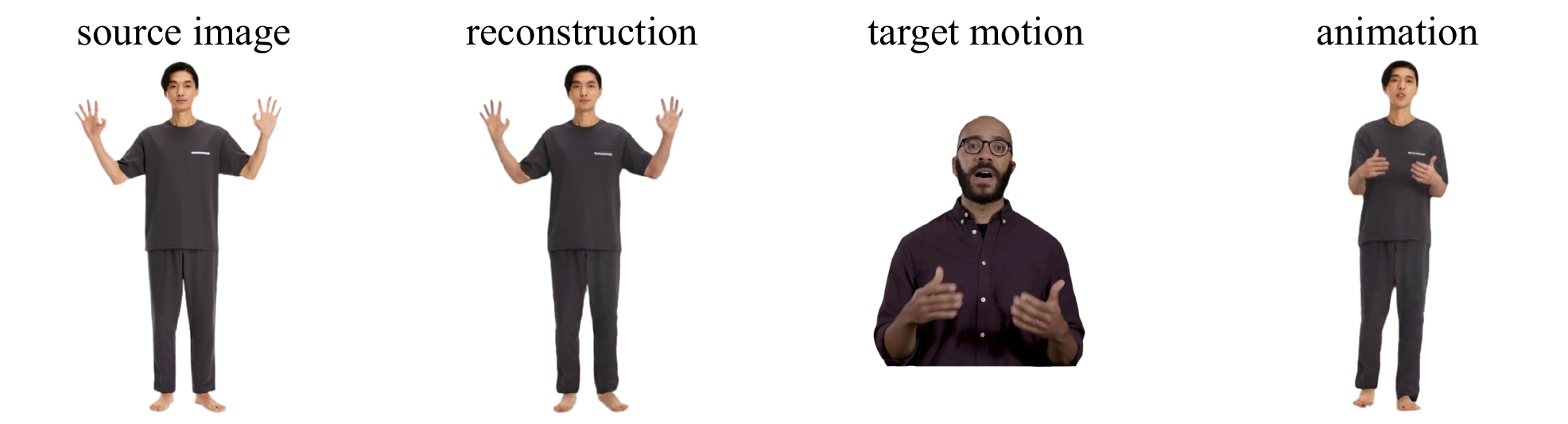}
   \caption{Full-Body reconstruction results. Our method successfully performs full-body reconstruction. Acquiring additional full-body data is expected to improve the results in such cases.}
   \label{supfig:show-whole}
\end{figure*}

\begin{figure*}[t]
  \centering
   \includegraphics[width=1.0\linewidth]{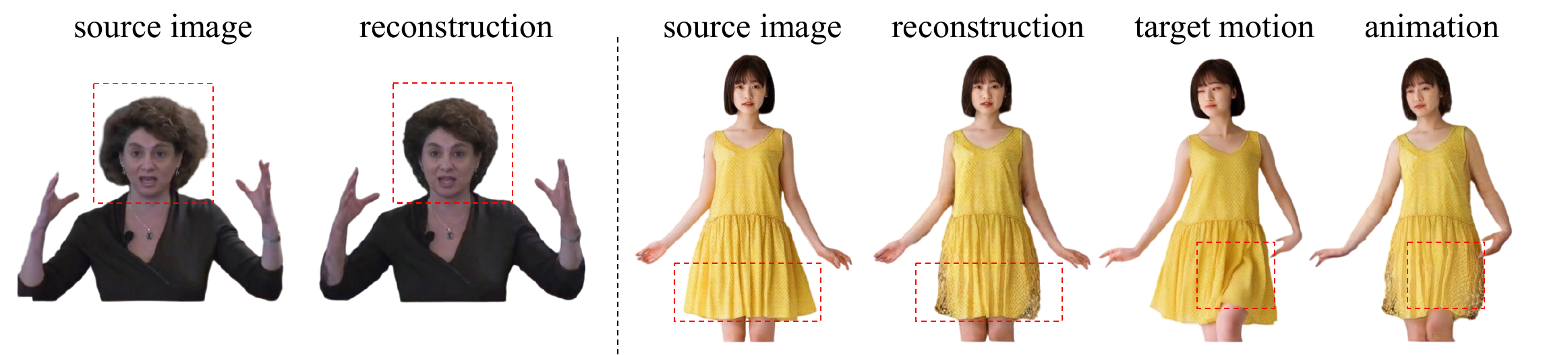}
   \caption{Visualization of failure cases. Our method exhibits limitations when handling fluffy hair, loose clothing, and flowing folds.}
   \label{supfig:failure-cases}
\end{figure*}

\begin{figure*}[t]
  \centering
   \includegraphics[width=1.0\linewidth]{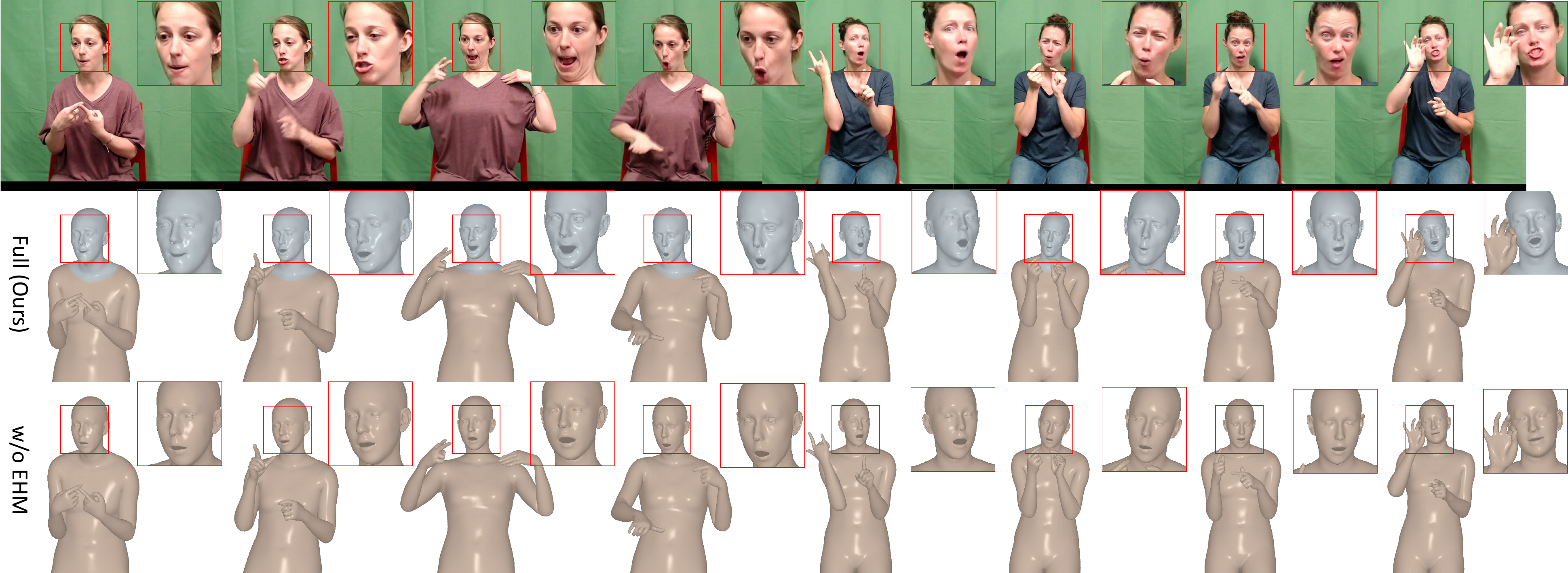}
   \caption{Visual results of our EHM tracking method. Without EHM, the model can only capture basic mouth-opening and -closing movements, whereas our method accurately tracks subtle facial expressions. Additionally, our approach successfully captures complex and detailed hand gestures.}
   \label{supfig:show-ehm}
\end{figure*}

\subsection{EHM tracking}
In the main paper Sec.~3.1, we briefly introduce the tracking of the template model's parameters using keypoint loss, omitting some details for clarity. However, the actual process is more complex. Here, we provide a more detailed description of our tracking method. Given images of a human’s upper body, we first estimate the keypoints $K_b$ using DWPose \cite {dwpose}. Based on these detected keypoints, we crop the head and hand regions. During the cropping process, we also record the affine transformation matrices $A_f$ and $A_h$. For the cropped hand image, we use HaMeR \cite{hamer} to estimate the parameters $Z_h=(\beta_h, \theta_h)$ for both hands. For the upper body, we estimate the SMPLX \cite{SMPL-X:2019} parameters $Z_b=(\beta_b,~\theta_{b})$ using PIXIE \cite{PIXIE:2021}.

\noindent \textbf{Face Tracking.} Based on the cropped head image, we estimate three sets of keypoints $K^1_f$, $K^2_f$ and $K^3_f$ using FaceAlignment \cite{facealignment}, MediaPipe \cite{lugaresi2019mediapipe}, InsightFace \cite{insightface-landmark}, where we only use the mouth keypoints from $K^3_f$. We also perform a rough estimation of the FLAME \cite{FLAME:SiggraphAsia2017} parameters $Z_f=(\beta_f,~\psi_f,~\theta_{f})$ using Teaser \cite{liu2025teaser}, where $\theta_{f}$ includes the jaw pose $\theta_{jaw}$, eye pose $\theta_{eye}$, neck pose $\theta_{neck}$ and a global pose $\theta_{fg}$. Then, we optimize the rotation $R$ and translation $T$ of the camera parameters, as well as $Z_f$, for 1000 iterations, with the loss function defined as follows:
\begin{equation}
    \begin{split}
    \mathcal{L}_{face-track}&=\sum_{i}^3\lambda^f_{k,i}\mathcal{L}_1(K_f^i,\hat{K_f^i})+
    \\&\lambda^f_{smo}\mathcal{L}_{smo}(Z_f,R,T)+\lambda^f_{reg}\mathcal{L}_{reg}(Z_f).
    \end{split}
\end{equation}
Here, $\mathcal{L}_{smo}$ and $\mathcal{L}_{reg}$ represent the smoothness loss between adjacent frames and the regularization loss (constraining parameters toward zero), respectively. Next, we optimize the eye pose $\theta_{eye}$ for 500 iterations using keypoint loss and smoothness loss, focusing on the eye keypoints.

Additionally, since the FLAME model does not include a mouth interior mesh, we follow \cite{qian2024gaussianavatars} to incorporate teeth into FLAME. The upper and lower teeth meshes are initialized accordingly, with their poses driven by the neck and jaw joints, respectively.

\noindent \textbf{Body Tracking.} After head tracking, we replace the head part of SMPLX with the neutral-pose expressive FLAME model $M_f(\beta_f,\psi_f,\theta_{jaw},\theta_{eye})$ (with zero global and neck pose) to obtain EHM, as described in main paper Eq.~1. We then optimize the body parameters using not only 2D keypoint loss but also a 3D guidance loss from the tracked FLAME model $M_f(Z_f)$ and the tracked MANO model $M_h(Z_h)$. Since these tracked models align with their respective cropped image regions, they serve as accurate guidance. By applying the recorded affine transformation, we convert their vertices form local to global space and compute the error between EHM’s head and hand vertices and these references, enhancing pose optimization and alignment accuracy. The following loss function is used to optimize $Z_b$, $\beta_f$, $\theta_h$ as well as camera parameters $R$ and $T$:

\begin{equation}
    \begin{split}
\mathcal{L}_{body-track}&=\lambda^b_k\mathcal{L}_1(K_b,\hat{K_b})+\lambda^b_{reg}\mathcal{L}_{reg}+\\&
\lambda^b_{smo}\mathcal{L}_{smo}(Z_b,\theta_h,R,T)(Z_b,\theta_h)+\\&\lambda^b_{3d}\mathcal{L}_{3d}(M_{ehm},A_f^{-1}M_f(Z_f),A_h^{-1}M_h(Z_h))+
\\&\lambda^b_{prior}\mathcal{L}_{prior}(Z_b ),
    \end{split}
\end{equation}

where, $\mathcal{L}_{prior}$ represents the constraint loss applied to the pose parameters using VPoser \cite{SMPL-X:2019}, enforcing a prior distribution.

\section{More Results}
\label{spsec:more-result}
\subsection{Novel views synthesis}
Reconstructing a 3D upper-body avatar from a single image is an ill-posed problem. However, GUAVA learns the animation of diverse subjects from different viewpoints, enabling it to generalize and infer certain 3D information. As a result, the reconstructed avatar not only supports animation but also enables novel view synthesis. \cref{supfig:show-novel-views} presents our results, demonstrating high multi-view consistency while preserving the subject’s identity. Additionally, synthesized novel views exhibit high-quality rendering with fine texture details.

We also demonstrate novel view synthesis from extreme angles, as shown in \cref{supfig:show-extreme-novel}. Since our training dataset includes only front views, back rendering is suboptimal. Training on 360° datasets, as IDOL \cite{idol} could help.

\subsection{Full-body}
Our method is also applicable to full-body settings as \cref{supfig:show-whole}. However, due to few full-body data in our dataset, adding more data could improve generalization.

\subsection{Failure cases}
As shown in \cref{supfig:failure-cases}, our method has limitations with fluffy hair, loose clothing, and flowing folds — all calling for future improvements.

\subsection{EHM tracking results}
Although we have demonstrated the improvement in model performance with the EHM model from both qualitative and quantitative perspectives in the main paper Sec.~4.3, to provide a more intuitive comparison, we further present the visual tracking results in \cref{supfig:show-ehm}. It is important to note that "w/o EHM" refers to tracking with SMPLX, which still uses our designed tracking framework, but without the FLAME integration step. From the results, it is clear that SMPLX can only roughly capture mouth movements, while EHM captures detailed facial expressions. Furthermore, our designed tracking framework not only captures accurate facial expressions but also tracks fine gestures, including finger movements, with high precision.

\section{More discussion}

\subsection{EHM vs SMPLX}
As discussed in the main paper Sec. 2.1, although SMPLX  \cite{SMPL-X:2019} integrates SMPL \cite{SMPL:2015} and FLAME \cite{FLAME:SiggraphAsia2017}, its expression space is newly trained on full-body scans, which may overlook fine facial details. This results in SMPLX having less expressive facial expressiveness compared to FLAME, a limitation also noted in ExAvatar (Sec. 3.1) \cite{moon2024exavatar}. EHM's main contribution is to improve this facial expressiveness.

\subsection{Ethical considerations}
\label{spsec:more-discuss}
\noindent  The generalization of 3D human avatar reconstruction technology raises several potential ethical concerns. First, unauthorized data collection and processing could lead to privacy violations, particularly with sensitive personal information like facial features and body shape. Second, this technology could be misused to create deepfake content, leading to identity theft, fraud, and other illegal activities. Additionally, the rapid reconstruction and real-time animation could be exploited to spread misinformation or engage in online harassment. Therefore, strict adherence to data protection regulations, ensuring informed consent, and taking measures to prevent misuse are essential. Transparency and traceability of technology should also be prioritized to build public trust and minimize potential risks.


\end{document}